\documentclass[runningheads]{llncs}

\usepackage{eccv}

\usepackage{eccvabbrv}
\usepackage{comment}
\usepackage{graphicx}
\usepackage{graphicx}
\usepackage{booktabs}
\usepackage{multirow}
\usepackage{adjustbox}
\usepackage{colortbl}
\usepackage{makecell}
\usepackage{wrapfig}
\usepackage[accsupp]{axessibility}  % Improves PDF readability for those with disabilities.

\usepackage{floatrow}
\newfloatcommand{capbtabbox}{table}[][\FBwidth]

\usepackage{hyperref}

\usepackage{orcidlink}

\begin{document}

% \title{Learning to Localization Actions in Instructional Videos Through Narrations} 

% \title{Action Localization in Instructional Videos with LLMs-based Pseudo Labeling} 

% \title{Learning to %Localization 
% Localize Actions in Instructional Videos with %LLMs=
% LLM-Reinforced Pseudo Labeling} 

\title{Learning to Localize Actions in Instructional Videos with LLM-Based Multi-Pathway Text-Video Alignment}

% as an abbreviated paper title here. If not, comment out.
\titlerunning{Localizing Actions in Videos with LLM-Based Multi-Pathway Alignment}

% Include the authors' OCRID for the camera-ready version, if possible.
\author{Yuxiao Chen\inst{1} \and
Kai Li \inst{2} \and
Wentao Bao\inst{4} \and Deep Patel \inst{3} \and \\ Yu Kong \inst{4} \and Martin Renqiang Min \inst{3} \and Dimitris N. Metaxas \inst{1}}

\authorrunning{Y.~Chen et al.}

\institute{Rutgers University \email{\{yc984, dnm\}@cs.rutgers.edu}, \and
Meta \email{\{li.gml.kai@gmail.com\}} \and
NEC Labs America-Princeton \email{\{dpatel, renqiang\}@nec-labs.com}
 \and
Michigan State University
\email{\{baowenta, yukong\}@msu.edu}}

% \institute{\textsuperscript{1} Rutgers University, \textsuperscript{2} NEC Labs America-Princeton, \textsuperscript{3} Michigan State University.
% }

\maketitle

\begin{abstract}
Learning to localize temporal boundaries of procedure steps in instructional videos is challenging due to the limited availability of annotated large-scale training videos. Recent works focus on learning the cross-modal alignment between video segments and ASR-transcripted narration texts through contrastive learning. However, these methods fail to account for the alignment noise, \ie, irrelevant narrations to the instructional task in videos and unreliable timestamps in narrations. To address these challenges, this work proposes a novel training framework. Motivated by the strong capabilities of Large Language Models (LLMs) in procedure understanding and text summarization, we first apply an LLM to filter out task-irrelevant information and summarize task-related procedure steps (LLM-steps) from narrations. To further generate reliable pseudo-matching between the LLM-steps and the video for training, we propose the Multi-Pathway Text-Video Alignment (MPTVA) strategy. The key idea is to measure alignment between LLM-steps and videos via multiple pathways, including: (1) step-narration-video alignment using narration timestamps, (2) direct step-to-video alignment based on their long-term semantic similarity, and (3) direct step-to-video alignment focusing on short-term fine-grained semantic similarity learned from general video domains. The results from different pathways are fused to generate reliable pseudo step-video matching. We conducted extensive experiments across various tasks and problem settings to evaluate our proposed method. Our approach surpasses state-of-the-art methods in three downstream tasks: procedure step grounding, step localization, and narration grounding by 5.9\%, 3.1\%, and 2.8\%.

\end{abstract}

\section{Introduction}
\label{sec:intro}

%introduction of instruction video and 

Instructional videos are a type of educational content presented in video format. They demonstrate the steps required to perform specific tasks, such as cooking, repairing cars, and applying makeup \cite{crosstask,coin,procel,howto100m}. These videos are invaluable resources for individuals seeking to learn new skills or knowledge in various domains. Instructional videos are in long duration since they aim to provide comprehensive visual guidance and instructions on task execution. Therefore, developing an automated system that can temporally localize procedural steps within instructional videos is crucial to facilitate the accessibility and utility of these educational resources. 

Previous works addressed the task of localizing procedure steps either in a fully supervised~\cite{activitynet,coin,cao2022locvtp,lea2016temporal} or weakly supervised ~\cite{setsup,crosstask,chang2019d3tw} manners. These methods train models with annotated temporal boundaries or sequence orders of procedure steps. Although these methods have achieved notable performance, they require massive training videos with annotations, which are expensive and labor-intensive to collect. Recent methods \cite{vina,tan,stepformer,dwsa} have explored training models using narrated instructional videos. Specifically, they train models to learn the cross-modal alignment between the text narrations and video segments through contrastive learning~\cite{milnce}, where the temporally overlapped video segments and narrations are treated as positive pairs.

\begin{figure}[t]
\centering
\includegraphics[width=\linewidth]{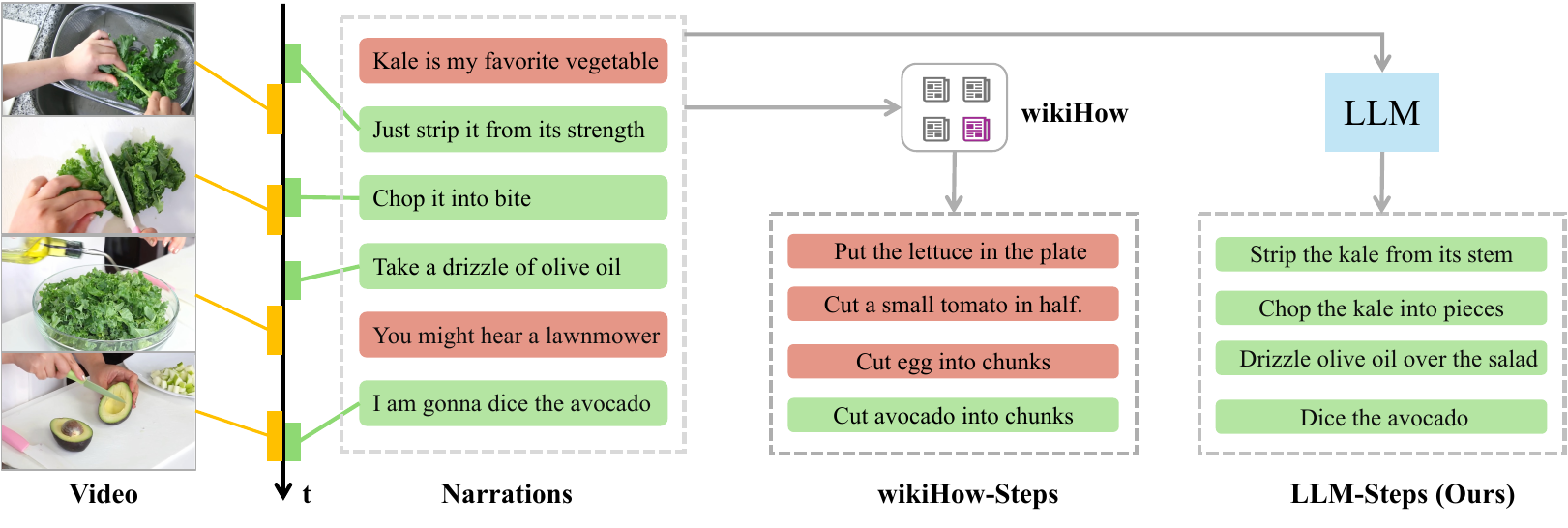}
  \caption{Illustration of different text information in narrated instruction videos. The orange and green bars along the time (t) axis denote the temporal boundaries of procedure steps and timestamps of narrations, respectively. Sentences highlighted in green indicate task-relevant information, while those in red are task-irrelevant.}
\label{fig:fig_1}
\vspace{-5pt}
\end{figure}

However, this approach is suboptimal since narrations are \textit{noisily} aligned with video contents~\cite{tan}. The noise mainly originates from two aspects. Firstly, some narrations may describe the information that is irrelevant to the task of the video. For instance, as illustrated in Figure \ref{fig:fig_1}, narrations highlighted in red boxes discuss the demonstrator's interest in a food or background sounds, which are not related to the task of making a salad. Furthermore, the temporal correspondence between video frames and narrations derived from narrations' timestamps may not be consistently reliable since demonstrators may introduce steps before or after executing them. As illustrated in Figure~\ref{fig:fig_1}, the demonstrator introduces the procedure ``take a drizzle of olive oil'' before actually performing it.

Previous methods~\cite{distsup,zhou2023procedure,vina} propose to mitigate the first problem by leveraging the procedure steps defined by the human-constructed knowledge base wikiHow~\cite{wikihow}, denoted as wikiHow-steps, to supplement or replace narrations. However, wikiHow-steps may not always align with the actions demonstrated in the videos, as a task can be executed in ways that are not outlined in the knowledge base. For example, as shown in Figure~\ref{fig:fig_1}, the wikiHow-step ``cutting tomato'' is absent in the videos, while some steps performed in the videos, such as ``take a drizzle of olive oil'', are not included in the knowledge bases.  Additionally,  previous stuies~\cite{tan,vina} employs self-training technologies to tackle the issue of temporal misalignment. They first train models with supervision signals derived from narration timestamps, and then use the trained models to filter out unmatched narrations and establish new pseudo-alignment between narrations and videos. However, the effectiveness of these methods may be constrained, since the models employed to filter noises are trained on noisy narrations. 

To address the above challenges, we propose a novel framework for training procedure steps localization models on narrated instructional videos. Figure~\ref{fig:method_overview} gives an overview of our proposed methods. Initially, we utilize a Large Language Model (LLM)~\cite{brown2020language,llamav1,llamav2,t5} to filter out task-irrelevant information and summarize the procedure steps from narrations. LLMs have demonstrated strong capabilities in understanding procedures \cite{driess2023palm,ahn2022can,antgpt}, thus serving as a knowledge base for instructional tasks. Additionally, LLMs possess robust text summarization capabilities~\cite{pu2023summarization,van2023clinical}. Therefore, by using LLM, we can extract \textit{video-specific} task-related procedure steps, denoted as LLM-steps. Compared with wikiHow-steps, which is designed to provide a general introduction to a task, our video-specific LLM-steps can align more closely with the video contents, as shown in Figure~\ref{fig:fig_1}.

Furthermore, we propose the Multi-Pathway Text-Video Alignment (MPTVA) strategy to generate reliable pseudo-matching between videos and LLM-steps. Our key insight is to measure the alignment between LLM-steps and video segments using different pathways, each of which captures their relationships from different aspects and provides complementary information. Consequently, combining those information can effectively filter out noise. Specifically, we leverage three distinct alignment pathways. The first pathway is designed to leverage the temporal correspondence derived from the timestamps of narrations. It first identifies narrations semantically similar to the LLM-steps and subsequently matches each LLM-step with video segments that fall within the timestamps of their semantically similar narrations. To mitigate the impact of timestamp noise, we further directly align LLM-steps with videos using a video-text alignment model pre-trained on long-duration instructional video data~\cite{vina}. This pathway measures the alignment based on long-term global text-video semantic similarity. Additionally, we directly align LLM-steps with video segments utilizing a video-text foundation model~\cite{internvideo} pre-trained on short-duration video-text datasets from various domains. This pathway not only captures fine-grained, short-term relationships but also incorporates broader knowledge learned from various video domains. We combine the alignment scores from the three pathways using mean-pooling, and then generate pseudo-labels from the obtained alignment score. It effectively leverages the mutual agreement among different pathways, which has been demonstrated to be effective in filtering noise for pseudo-labeling~\cite{tan}.

We conducted extensive experiments across various tasks and problem settings to evaluate our proposed method. Our approach significantly outperforms state-of-the-art methods in three downstream tasks: procedure step grounding, action step localization, and narration grounding. Notably, our method exhibits improvements over previous state-of-the-art methods by 5.9\%,  2.8\%, and 3.1\% for procedure step grounding, narration grounding, and action step localization. We also draw the following key observations: (1) Models trained with our LLM-steps significantly outperform those trained with wikiHow-steps and a combination of wikiHow-steps and narrations by 10.7\% and 6\%, demonstrating the superior effectiveness of our LLM-steps. (2) When employing the supervision generated by our proposed MPTVA, models trained with narrations outperform those trained with wikiHow-steps. This underscores the limitations of using wikiHow-steps for training (3) Our ablation study reveals that applying all three alignment pathways improves the performance by 4.2\% and 5.4\% compared to the best model utilizing a single pathway, and by 3\% and 4.5\% compared to the best model utilizing two pathways. This indicates that the complementary information contributed by each pathway results in more reliable pseudo-labels.

\section{Related Work}
\textbf{Step Localization in Instructional Videos} Previous studies on step localization in instructional videos can be classified into fully-supervised~\cite{activitynet,coin,cao2022locvtp,lea2016temporal} and weakly-supervised~\cite{setsup,crosstask,chang2019d3tw,wsag} methods. Fully-supervised methods train models to directly predict the annotated boundaries of procedure steps. However, the main limitation of these approaches is the high cost to collect temporal annotations. On the other hand, weakly supervised methods only require the order or occurrence information of procedure steps for training, thus reducing the cost of annotation. However, collecting such annotations remains costly, as it requires annotators to watch entire videos. Recent work ~\cite{dwsa,stepformer,vina,tan} attempts to eliminate labeling efforts by using narrated instructional videos. Their main focus is on how to leverage the noisy supervision signals provided by narrations. For example, some works~\cite{dwsa,stepformer} proposed learning an optimal sequence-to-sequence alignment between narrations and video segments by utilizing DTW~\cite{dtw,dropdtw}. However, these methods are not suited for the downstream tasks, such as step grounding, where the order of steps is unknown. Our method is more related to VINA~\cite{vina} and TAN~\cite{tan}. TAN proposed filtering out noisy alignments through mutual agreement between two models with distinct architectures. VINA extracted task-specific procedure steps from the WikiHow knowledge base to obtain auxiliary clean text descriptions. It aslo explored generating pseudo-alignments between procedure steps and videos using pre-trained long-term video-text alignment models. In contrast, our method leverages LLMs to extract video-specific task procedure steps, which align more closely with video contents than task-specific procedures used by VINA.  Furthermore, our proposed MPTVA leverages a broader range of information which is extracted from various alignment paths and models pre-trained on diverse datasets, to generate pseudo-labels.

\noindent\textbf{Learning from Noisily-Aligned Visual-Text Data.} Recent studies have achieved remarkable progress in extracting knowledge from large-scale, weakly aligned visual-text datasets. In the field of images~\cite{clip,jia2021scaling,hiclip,fdt}, it has been demonstrated that features with strong generalization abilities to various downstream tasks can be learned from large-scale weakly aligned image-text datasets through contrastive learning. In the video domain, this method has been extended to enhance video understanding. Various pretraining tasks~\cite{k700,kinetics,bain2021frozen,internvideo,clipvip,zhao2024videoprism,chen2022hierarchically} have been proposed to capture dynamic information in videos, and extensive video-text datasets are constructed to train the model. In the domain of instructional video analysis, researchers~\cite{milnce,luo2020univl,zhou2023procedure,distsup,xu2021videoclip} mainly focus on learning feature representations from the HowTo100M dataset~\cite{howto100m}. They mainly utilize narrations and video frames that are temporally overlapped as sources for contrastive learning or masked-based modeling. In this work, we leverage knowledge learned by these models from various data sources and pertaining tasks to measure the correspondence between video content and LLM-steps. We then combine the results to generate pseudo-alignment between LLM-steps and videos.

\noindent\textbf{Training Data Generation Using LLMs} In order to reduce the cost of data collection, recent studies have explored leveraging LLMs to automatically generate training data. In the NLP domain~\cite{chatgptann,llmann,llmsent}, LLMs are employed to generate annotations for various tasks such as relevance detection, topic classification, and sentiment analysis. In the image-text area, LLMs are employed to generate question-answer text pairs from image captions for instruction tuning~\cite{llava,instructblip,liu2023improved,chen2023visual}. In video datasets, LLMs are primarily used to enrich the text descriptions of videos. For instance, MAXI~\cite{maxi} leverages LLMs to generate textual descriptions of videos from their category names for training. InternVideo~\cite{internvideo} utilizes LLMs to generate the action-level descriptions of videos from their frame-level captions. HowtoCap~\cite{howtocap} leveraged LLMs to enhance the detail in narrations of instructional videos. By contrast, our approach apply the procedure understanding and text summarization capabilities of LLMs to filter out irrelevant information and summarize the task-related procedural steps from narrations.

\section{Method}

\subsection{Problem Introduction}
\label{sec:problem}
In this work, we have a set of narrated instruction videos for training. A video $\mathcal{V}$ consists of a sequence of non-overlapping segments $\{v_1, v_2,..., v_T\}$, where $v_i$ is the $i$-th segment and $T$ is the number of video segments. The narration sentences $\mathcal{N} = \{n_1, n_2, ...,n_K\}$ of the video, where $K$ is the number of narrations, are transcribed from the audio by using the ASR system. Each narration $n_i$ is associated with the start and end timestamps within the video. We aim to use these videos and narrations to train a model $\Phi$ that takes a video and a sequence of sentences as inputs, and produces an alignment measurement
between each video segment and sentence. The trained $\Phi$ is deployed to identify the temporal regions of procedural steps within the videos. 

A straightforward way to solve the problem is to assume $\mathcal{N}$ are matched with the video segments within their timestamps, while unmatched with others. Specifically, $\Phi$ takes $\mathcal{V}$ and $\mathcal{N}$ as inputs to predict a similarity matrix between $\mathcal{V}$ and $\mathcal{N}$:
\begin{equation}
    \mathbf{\hat{A}} = \phi(\mathcal{N}, \mathcal{V}) \in \mathbb{R}^{K \times T}
\end{equation}
The model $\Phi$ is trained to maximize the similarity between matched video segments and narrations, while decrease those between unmatched ones. This is achieved by minimizing the MIL-NCE loss~\cite{milnce}:
\begin{equation}
    \label{eq:mil-nce}
    \mathcal{L}(\mathbf{Y}^{NV}, \mathbf{\hat{A}}) = - \frac{1}{K} \sum_{k=1}^{K} \log \left( \frac{\sum_{t=1}^{T} {Y}_{k,t}^{NV} \exp({\hat{A}}_{k,t} / \eta)}{\sum_{t=1}^{T} \exp({\hat{A}}_{k,t} / \eta)} \right),
\end{equation} where $\mathbf{Y}^{NV}$ is the \textit{pseudo matching matrix} between $\mathcal{N}$ and $\mathcal{V}$ constructed by the timestamps of narrations. ${Y}_{k,t}^{NV}$ is the element at the $k$-th row and $t$-th column. It is equal to one if the $v_t$ is within the timestamps of $n_k$, and zero otherwise. $\eta$ is the softmax temperature parameter. 

However, the solution is sub-optimal, since $n_i$ might not be relevant to any visual content of $V$, and the matching relationship encoded in $\mathbf{Y}^{NV}$ is unreliable, due to the temporal misalignment problem of timestamps.

\subsection{Proposed Framework}
\begin{figure}[t]
\centering
\includegraphics[width=\linewidth]{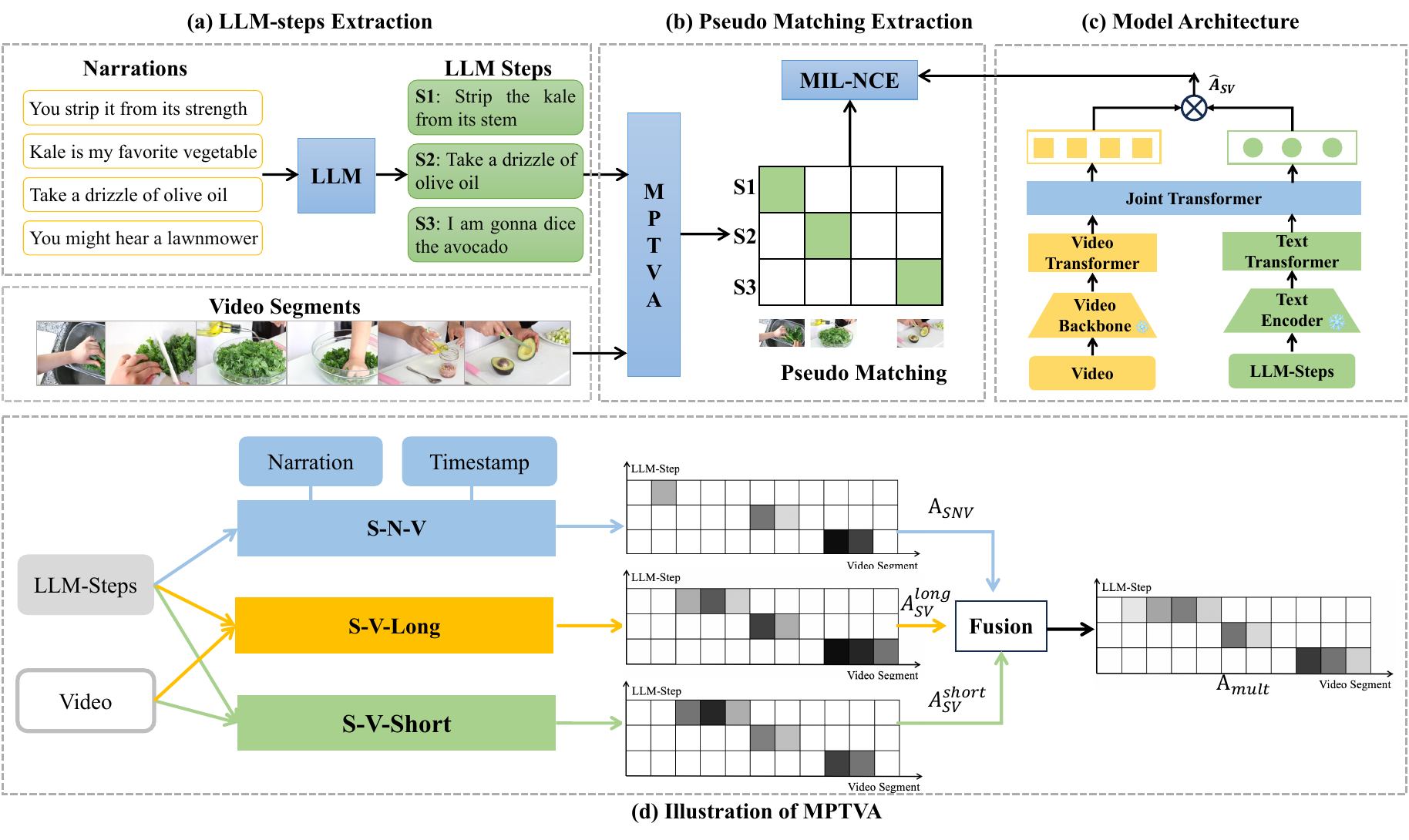}
  \caption{Overview of our proposed method. (a): We first use an LLM to summarize task-relevant LLM-steps from narrations. (b): We then extract the pseudo-matching between LLM-steps and video segments using our proposed MPTVA. (c): The extracted pseudo-alignments are used as supervision to train the model to minimize the MIL-NCE loss. (d): The illustration of the proposed MPTVA strategy for pseudo-label generation.}
\label{fig:method_overview}
\end{figure}

Figure \ref{fig:method_overview} gives an overview of our proposed methods. We first apply an LLM to process narrations $\mathcal{N}$, filtering out task-unrelated information and summarizing task-relevant procedure steps $\mathcal{S}$. Subsequently, we extract the pseudo matching matrix $\mathbf{Y}^{SV}$ between $\mathcal{S}$ and $\mathcal{V}$ by exploring multiple alignment pathways. The pseudo-alignments $\mathbf{Y}^{SV}$ are used as supervision to train the model $\Phi$ to learn the alignment between $\mathcal{S}$ and $\mathcal{V}$.

\subsubsection{Extracting Procedure Steps Using LLM.}
\label{sec:llm}
Recent studies demonstrate that LLMs possess a strong capability for text summarization~\cite{van2023clinical,pu2023summarization} and understanding task procedures~\cite{driess2023palm,ahn2022can,antgpt}.  Motivated by this, we propose using LLMs to filter video-irrelevant information and extract task-related procedural steps from narrations. To this end,  we design the following prompt $\mathbf{P}$, which instructs the LLM model to (1) summarize the procedure steps relevant to the task shown in the video and (2) filter out irrelevant information, such as colloquial sentences:

\textit{``I will give you the text narrations extracted from an instruction video. Your task is to summarize the procedure steps that are relevant to the task of the video from the inputs. Please filter colloquial sentences in the speech.''}

This prompt, along with the narrations $\mathcal{N}$, is concatenated and then fed into an LLM model, resulting in the LLM-generated procedural steps $\mathcal{S}$ (denoted LLM-steps). The process is formulated as follows:
\begin{equation}
\mathcal{S} = LLM([\mathbf{P}; \mathcal{N}]).% \quad
% \mathcal{S} = \{s_1, s_2 ..., s_L\}
\end{equation}
We denote $\mathcal{S} = \{s_1, s_2 ..., s_L\}$ where $s_i$ is the $i$-th out of $L$ LLM-steps. %, and $L$ is the number of LLM-steps.

Figure~\ref{fig:fig_1} shows examples of obtained LLM-steps. We can see that (1) $\mathcal{S}$ are more descriptive than $\mathcal{N}$, and (2) most of the information unrelated to the video task contained in $\mathcal{N}$, such as ``You might hear a lawnmower'', have been filtered out in $\mathcal{S}$. In addition, we observe $\mathcal{S}$ has retained most of the procedure information related to the task, and can align more contents of the videos than the wikiHow-steps. which are widely in the previous work~\cite{vina,distsup,zhou2023procedure}. Please refer to the supplementary material for more examples.

\subsubsection{Multi-Pathway Text-Video Alignment} The obtained LLM-steps are not accompanied by annotations indicating temporal alignment with videos. To tackle this challenge, we propose the Multi-Pathway Text-Video Alignment (MPTVA) strategy to generate reliable pseudo-alignment between LLM-steps and video segments for training. The key idea is to measure the alignment between $\mathcal{S}$ and $\mathcal{V}$ in different pathways, each of which provides complementary information to others. The results from different pathways are then fused to extract the pseudo-matching relationship.

Specifically, we measure their alignment using three different pathways. These include: (1) aligning $\mathcal{S}$ with narrations $\mathcal{N}$ and then to $\mathcal{V}$ (S-N-V), (2) directly aligning $S$ to $\mathcal{V}$ using the long-term video-text alignment models pre-trained in the instructional video domain ({S-V-long}), and (3) direct alignment using a video-text foundation model pre-trained on short-term video-text datasets from various domains (S-V-short). The three pathways capture the relationship between $\mathcal{S}$ and $\mathcal{V}$ based on different information, including the timestamps of narrations, long-term global text-video semantic similarity, and short-term fine-grained text-video semantic similarity.

The S-N-V path first identifies semantically similar narrations of $\mathcal{S}$, and then associates $\mathcal{S}$ with the video frames that are within the timestamps of these narrations. Formally, the semantic similarity distributions between the $\mathcal{S}$ and $\mathcal{N}$ of a video are computed as below: 
\begin{equation}
    \label{eq:sn_att}
    \mathbf{A}_{SN} = \mathrm{Softmax}\left(\frac{E_t(\mathcal{S}) \cdot  {E_t(\mathcal{N})^\top }}{\tau}\right) 
\end{equation}
where $\mathbf{A}_{SN}\in \mathbb{R} ^ {L \times K}$, $E_t(\mathcal{S}) \in \mathbb{R} ^ {L \times D}$, and $E_t(\mathcal{N}) \in \mathbb{R} ^ {K \times D}$. Note that $E_t(\mathcal{S})$ and $E_t(\mathcal{N})$ are the L2 normalized embeddings of the $\mathcal{S}$ and $\mathcal{N}$ extracted using a pre-trained text encoder $E_t$, respectively. % $\cdot$ denotes the matrix multiplication operation. 
$\tau$ is the temperature.
Then, the alignment score between LLM steps and video frames can be obtained as follows:
\begin{equation}
    \label{eq:snv}
    \mathbf{A}_{SNV} = \mathbf{A}_{SN} \cdot \mathbf{Y}^{NV} \in \mathbb{R} ^ {L \times T}
\end{equation}
$\mathbf{Y}^{NV}$ is the pseudo matching matrix constructed based on the timestamps of narrations, as introduced in Equation \ref{eq:mil-nce}.

Since the temporal alignment derived from timestamps can be noisy, we incorporate additional information by directly aligning LLM steps and video segments. This is achieved by using a \textit{long-term} video-text alignment model pre-trained using the instructional video data. 

Specifically, we first use the text encoder $E_t^L$ and video encoder $E_v^L$ of the model to extract the embeddings of the $\mathcal{S}$ and $\mathcal{V}$, respectively. The cosine similarity between each LLM-step and video segment is calculated, resulting in the long-term step-to-video alignment score matrix $\mathbf{A}_{SV}^{long}$: %. It is formulated as bellow:
\begin{equation}
\mathbf{A}_{SV}^{long} = E_t^L(\mathcal{S})\cdot E_v^L(\mathcal{V})^\top.
\label{eq:asv_shot}
\end{equation}

We also obtained additional direct alignment between LLM steps and videos, denoted as $\mathbf{A}_{SV}^{short}$, using the video foundation models pre-trained on general short-term video-text data. $\mathbf{A}_{SV}^{short}$ is obtained follow the same way as Equation \ref{eq:asv_shot}, but use the text encoders $E_t^S$ and $E_v^S$ of the video foundation model.  This pathway complements the alignment by focusing on short-term, fine-grained information, such as human-object interactions. Furthermore, this pathway introduces knowledge learned from a broader domain. 

The multiple pathways alignment score matrix $\mathbf{A}_{mult}$ is obtained by mean-pooling $\mathbf{A}_{SNV}$, $\mathbf{A}_{SV}^{long}$, and $\mathbf{A}_{SV}^{short}$.

\textbf{Discussion.} Previous works typically generate pseudo-labels by focusing on a single path, either $\mathbf{A}_{SNV}$ or $\mathbf{A}_{SV}^{\text{long}}$. It potentially results in labels that are affected by the noise present in each pathway. In contrast, our method leverages information from multiple paths, thereby enabling the generation of more reliable pseudo-alignments. Furthermore, our proposed $\mathbf{A}_{SV}^{short}$ introduces additional knowledge beyond the instructional video domain. Our method shares the same idea with TAN \cite{tan} that uses the mutual agreement between different models to filter noise in pseudo-labels. However, our approach leverages a broader range of complementary information through multiple alignment pathways and models pre-trained on a variety of datasets. 

%In contrast, TAN relies on two models that are trained on instructional videos using noisy supervision derived from timestamps. 

\subsubsection{Learning LLM-step Video Alignment} 
Given $\mathbf{A}_{mult}$, we construct the pseudo-matching matrix $\mathbf{Y}^{SV}$ as follows: For an LLM step $s_i$, we first identify the video segment $v_k$ that has the highest alignment score with $s_i$. $v_k$ is treated as the matched video segment for $s_i$, and thus we set $\mathbf{Y}^{SV}{(i,k)}$ to one. Additionally, for any video segment $v_j$, if its temporal distance from $v_k$ (defined as $|j-k|$) falls within a predetermined window size $W$, it is also labeled as a matched segment to $s_i$, leading us to mark $\mathbf{Y}^{SV}{(i,j)}$ as ones accordingly. To ensure the reliability of pseudo-labels, we exclude LLM steps for which the highest score in $\mathbf{A}_{mult}$ falls below a specified threshold $\gamma$.

We use $\mathbf{Y}^{SV}$ as the supervision to train $\Phi$ to learn the alignment between $\mathcal{S}$ and $\mathcal{V}$. The model $\Phi$ takes $\mathcal{S}$ and $\mathcal{V}$ as inputs, and output a similarity matrix $\mathbf{\hat{A}_{SV}}$ between $\mathcal{S}$ and $\mathcal{V}$:
\begin{equation}
    \mathbf{\hat{A}_{SV}} = \Phi(\mathcal{S}, \mathcal{V}),
\end{equation}
where $\mathbf{\hat{A}_{SV}} \in \mathbb{R}^{L\times T}$. The model $\Phi$ is trained to pull matched video segments and LLM-steps closely while pushing way unmatched ones, by minimizing the MIL-NCE loss $\mathcal{L}(\mathbf{Y}^{SV}, \mathbf{\hat{A}_{SV}})$ shown in Equation \ref{eq:mil-nce}.

\subsection{Architecture}

Figure~\ref{fig:method_overview} illustrates the architecture of $\Phi$. It consists of the video backbone and text encoder, the unimodal Transformers encoder~\cite{selfatt} for video and text, and a joint-modal Transformer.
Given the input video $\mathcal{V}$ and LLM-steps $\mathcal{S}$, we first extract their embeddings by using the video backbone $f_{b}$ and text backbone $g_{b}$:

\begin{equation}
\mathbf{F}_V = f_{b}(\mathcal{V}), \quad \mathbf{F}_S = g_{b}(\mathcal{S}),
\end{equation}
where $\mathbf{F}_V\in \mathbb{R}^{T \times D}$ and $\mathbf{F}_S\in \mathbb{R}^{L \times D}$. 
The extracted video segments and text features from the backbone are subsequently fed into the model-specific unimodal Transformers, $f_{trs}$ for video and $g_{trs}$ for text, to refine the features by modeling the intra-modal relationships:
\begin{equation}
    \overline{\mathbf{F}}_V = f_{trs}(\mathbf{F}_V + \mathbf{F}_{\text{PE}}), \quad   
    \overline{\mathbf{F}}_S = g_{trs}(\mathbf{F}_S),
\end{equation}
where $\overline{\mathbf{F}}_V \in \mathbb{R}^{T \times D}$ and $\overline{\mathbf{F}}_S\in \mathbb{R}^{L \times D}$.
Note that we add positional embedding (PE), denoted as $\mathbf{F}_{\text{PE}}$, into the video segments to represent the time orders of video segments.  We do not use PE with LLM-steps since the input procedure steps may not have sequential orders on downstream tasks \cite{crosstask,vina}.

Finally, the joint Transformer $h_{trs}$ takes the sequential concatenation of $\hat{\mathbf{F}}_V$ and $\hat{\mathbf{F}}_S$ as inputs to further model the fine-grained cross-modal relationship between video segments and LLM-steps. It outputs the final feature representations of video segments $\hat{\mathbf{F}}_V$ and LLM-steps $\hat{\mathbf{F}}_S$.

\begin{equation}
    [\hat{\mathbf{F}}_V; \hat{\mathbf{F}}_S] = h_{trs}([\overline{\mathbf{F}}_V; \overline{\mathbf{F}}_S])
\end{equation}

The similarity matrix $\mathbf{\hat{A}}_{SV}$ between LLM-steps and video segments are obtained by calculating the cosine similarity between $\hat{\mathbf{F}}_V$ and $\hat{\mathbf{F}}_S$:
\begin{equation}
    \mathbf{\hat{A}}_{SV} = {\hat{\mathbf{F}}_S} \cdot \hat{\mathbf{F}}_V^\top.
\end{equation}

\section{Experiment}
\subsection{Dataset}
\textbf{HTM-370K (Training)} Following the previous work \cite{tan,vina} we train our model on the Food \& Entertaining subset of Howto100M datasets~\cite{howto100m}. It consists of about 370K videos collected from YouTube. 

\noindent\textbf{HT-Step (Evaluation)} This dataset~\cite{vina} consist of 600 videos selected from the Howto100M dataset~\cite{howto100m}. Each video is manually matched to an instructional article of the wikiHow dataset~\cite{wikihow}. The temporal boundaries of the procedure steps from the instructional articles are annotated if they are present in the videos. We use the dataset to evaluate the performance of models for \textit{procedure step grounding}. Following the previous work \cite{vina}, we report the R@1 metric. R@1 is defined as the ratio of successfully recalled steps to the total number of steps that occur, where a step is considered as successfully recalled if its most relevant video segment falls within the ground truth boundary.

\noindent\textbf{CrossTask (Evaluation)} This dataset~\cite{crosstask} has about 4.7K instructional videos, which can be divided into 65 related tasks and 18 primary tasks. Following the previous study~\cite{vina}, we use the dataset to evaluate the model for zero-shot  \textit{action step localization} task. Action step localization is to localize the temporal positions for each occurring action step. Compared to the procedure steps in the HT-Step dataset, the action steps represent more atomic actions, and their text descriptions are more abstract, containing fewer details. The metric is Average Recall@1 (Avg. R@1)~\cite{vina}. It's computed by first calculating the R@1 metric for each task and then averaging across tasks. We report the results averaging across 20 random sets, each containing 1850 videos from the primary tasks \cite{vina}.

\noindent\textbf{HTM-Align (Evaluation)} This dataset \cite{tan} contains 80 videos with annotated temporal alignment between ASR transcriptions and the video contents. It is used to evaluate our model for narration grounding. We report the R@1 metric following the common practice~\cite{tan,wsag}.

\subsection{Implementation Details}
Following the common practice~\cite{tan,vina}, we equally divide each input video into a sequence of 1-second non-overlapped segments and set the frame rate to 16 FPS. The pre-trained S3D video encoder, released by~\cite{milnce}, is used as the video backbone $f_b$ to extract a feature embedding from each segment. Following \cite{tan,vina}, $f_b$ is kept fixed during training. The text backbone $g_b$ is a Bag-of-Words (BoW) model based on Word2Vec embeddings \cite{milnce}. Following~\cite{vina}, we use the TAN* models, a variance of the TAN model~\cite{tan} that is pre-trained on long-term instructional videos, as $E_e^L$ and $E_e^L$. 
We also use the text encoder of TAN* as $E_t$. In addition, we set $E_t^S$ and $E_v^S$ as the pre-trained text and video encoders of the InternVideo-MM-L14 model~\cite{internvideo}. We use Llama2-7B~\cite{llamav2} to extract LLM-steps from narrations. We set the temperature hyperparameters $\eta$ and $\tau$ to 0.07. The model is trained on Nvidia A100 GPUs. We set the batch size to 32. The AdamW optimizer with an initial learning rate of $2 \times 10^{-4}$ is employed for training. The learning rate follows a cosine decay over 12 epochs. For further details, please refer to the supplementary material.

\subsection{Ablation Study}

\begin{table}[t]
\centering

\caption{Results for the step grounding and action step localization on the HT-Step and CrossTask datasets when using different pathways to generate pseudo-labels. "ZS" indicates applying the correspondent pre-trained video-text alignment models to perform the downstream tasks in a zero-shot manner. The symbol $\surd$ indicates that the respective pathways are used.} %"S$\rightarrow$N$\rightarrow$V" represents pseudo-labels based on alignment from steps to narration and then to videos. "N$\rightarrow$V" signifies pseudo-labels derived from alignment between narration and video}
\label{tbl:multi_pl}
\footnotesize
\setlength{\extrarowheight}{0mm}
\setlength{\tabcolsep}{0.5mm}
% \resizebox{0.7\linewidth}{!}{
\begin{tabular}{@{}ccccc@{}}
\toprule
\multicolumn{3}{c}{Pathway Type} & HT-Step              & CrossTask               \\ \cmidrule(r){1-3} \cmidrule(l){4-4} \cmidrule(l){5-5} 
{ S-N-V }    & { S-V-Long }  & { S-V-Short } & {  R@1 (\%) $\uparrow$\  } & { Avg. R@1(\%)$\uparrow$\ } \\ \midrule
         & ZS        &           & 30.3                 & 32.3                    \\
         &           & ZS        & 37.3                 & 39.7                    \\ \midrule
$\surd$  &           &           & 34.2                 & 36.4                    \\
         & $\surd$   &           & 35.0                 & 39.1                    \\
         &           & $\surd$   & 37.7                 & 41.6                    \\ \midrule
$\surd$  & $\surd$   &           & 37.9                 & 42.6                    \\
$\surd$  &           & $\surd$   & 38.0                 & 43.7                    \\
         & $\surd$   & $\surd$   & 38.9                 & {42.5}                  \\ \midrule
$\surd$  & $\surd$   & $\surd$   & \textbf{41.9}        & \textbf{47.0}           \\ \bottomrule
\end{tabular}
% }

\label{tbl:abl_pstype}
\end{table}
\subsubsection{Effectiveness of  Multi-Pathway Text-Video Alignment} We evaluate the effectiveness of our proposed MPTVA strategy by comparing the performances of models trained with labels generated through various pathways. The results are shown in Table~\ref{tbl:multi_pl}. It is observed that models trained exclusively with S-V-Short or S-V-Long labels significantly outperform the pre-trained short-term and long-term video-text alignment models, respectively. The results suggest that the models learn knowledge surpassing the capabilities of the alignment models used for pseudo-label extraction. We can see that the model trained with S-N-V underperforms those trained with S-V-Short or S-V-Long. This may be attributed to the noise present in the timestamps. Moreover, we can observe that using any combination of two pathways achieves better performance improvement than using only one pathway. More importantly, the best improvement is achieved when considering all three pathways. The results demonstrate that each pathway provides complementary information. By fusing these pathways, we can leverage this combined information, resulting in more reliable pseudo-labels.

\begin{table}[t]
\setlength{\extrarowheight}{0mm}
\setlength{\tabcolsep}{5mm}
\centering
\caption{Results for step grounding and action step localization on the HT-Step and CrossTask datasets when using different training data. ``N'', ``W'', and ``S'' denote narrations, wikiHow-steps, and LLM-steps, respectively. ``TS'' and ``MPTVA'' denote pseudo-labels generated from timestamps and our MPTVA strategy, respectively. }

\begin{tabular}{@{}ccc@{}}
\toprule
\multirow{2}{*}{Training Data} & HT-Step        & CrossTask            \\ \cmidrule(l){2-3} 
                               & R@1 $\uparrow$ & Avg. R@1 $\uparrow$\ \\ \midrule
N-TS                           & 30.3           & 32.3                 \\
N-MPTVA                        & 32.1           & 38.0                 \\
W-MPTVA                        & 31.2           & 37.8                 \\ \midrule
N-MPTVA+W-MPTVA                & 35.9           & 40.6                 \\ \midrule
S-MPTVA                        & 41.9           & 47.0                 \\
N-MPTVA + S-MPTVA              & \textbf{43.3}  & \textbf{47.9}        \\ \bottomrule
\end{tabular}
\label{tbl:abl_input}
\vspace{-10pt}
\end{table}

\subsubsection{Effectiveness of LLM-steps} To evaluate the effectiveness of LLM-steps, we compare the model trained with different text inputs, including narrations, procedure steps extracted from the wikiHow database following ~\cite{vina}, and our LLM-steps. The results are reported in Table~\ref{tbl:abl_input}.  It can be observed that the N-TS model, which takes narrations as input and employs timestamps for supervision, exhibits the lowest performance. Using the pseudo-labels generated by our proposed MPTVA for narrations (N-MPTVA) leads to improved performance over N-TS, showing that our proposed MPTVA strategy is also effective for noisy narrations. Notably, this model (N-MPTVA) surpasses the one (W-MPTVA) utilizing wikiHow-steps as input. The results indicate that, when using the same supervision (MPTVA), wikiHow-steps are less informative than narrations. It may be attributed to mismatches between instruction articles and videos. Additionally, it could be due to the fact that even the matched instruction articles only cover a small portion of the procedure steps depicted in the videos, considering that a task can be performed in multiple ways. In addition, we observe that using narration and wikiHow articles together, as previous work \cite{vina}, can enhance model performance. Nonetheless, training the models only with our LLM-steps outperforms this setting by 6\% and 6.4\% for step-grounding and action step localization tasks, respectively. These results underscore the effectiveness of LLM-steps in providing highly relevant and clean information for procedure step localization. In addition, we also note that incorporating the information of narration with LLM-steps can further boost the performance for 1.4\% and 0.9\%.

\begin{table}[t!]
\caption{Results of the models trained with the pseudo-labels that are generated with different filter thresholds $\gamma$ and window sizes $W$.}
\centering
\begin{tabular}{@{}ccccc@{}}
\toprule
\multirow{2}{*}{  $\gamma$  } & \multirow{2}{*}{  $W$  } & \multirow{2}{*}{\makecell{Valid Step\\Ratio}} & {  HT-Step  } & { CrossTask } \\ \cmidrule(l){4-4}  \cmidrule(l){5-5}
     &    &      & R@1 (\%)      & Avg. R@1      \\ \midrule
%0.55 & 5  & 55\% &               & 41.8          \\
0.60  & 5  & 37\% & 39.2          & 44.2          \\
0.65 & 5  & 15\% & 40.6          & 45.2          \\
0.70  & 5  & 3\%  & 35.4          & 41.2          \\ \midrule
0.65 & 1  & 15\% & 40.3          & 45.7          \\
0.65 & 2  & 15\% & \textbf{41.9} & \textbf{47.0} \\
0.65 & 5  & 15\% & 40.6          & 45.2          \\ \bottomrule
%0.65 & 10 & 15\% & 37.0          & 42.0          \\ 
\end{tabular}
\label{tbl:window_filter}
\vspace{-10pt}
\end{table}

%\input{table/gamma_winsize}
%\vspace{-10pt}

\subsubsection{Ablation of Filtering Threshold and Window Size for Pseudo-label Generation} We report the performance of models trained with pseudo-labels generated with different filtering thresholds ($\gamma$) and window sizes ($W$) in Table~\ref{tbl:window_filter}. 
We first fix the $W$ as 5 to assess the influence of $\gamma$. We can see that the best performance is observed when $\gamma$ is set to 0.65, and about 15\% LLM steps are selected for training. Performance decreases as $\gamma$ decreases from 0.65, likely due to the introduction of more noisy pseudo-labels. In addition, increasing  $\gamma$ from 0.65 significantly drops the performances, since a large percent of data (97\%) is excluded.  We then fix $\gamma$ and evaluate the impact of varying $W$. We can the best performance is achieved when $W$ is 2. Increasing $W$ beyond this point reduces performance, likely due to the mislabeling of irrelevant frames as matched. Additionally, decreasing $W$ from 2 to 1 results in a slight decrease in performance, as it causes some nearby relevant frames to be annotated as unmatched.  

\begin{table}[t]
    \centering
    \footnotesize
    \setlength{\tabcolsep}{0.1mm}
    \setlength{\extrarowheight}{0.5mm}
    \begin{tabular}{@{}ccc@{}}
        \toprule
        \multirow{2}{*}{Method}        & HTM-Align       & HT-Step        \\ \cmidrule(l){2-2} \cmidrule(l){3-3}
                                       & R@1 $\uparrow$\ & R@1 $\uparrow$ \\ \midrule
        CLIP  (ViT-B/32)~\cite{clip}              & 23.4            & -              \\
        MIL-NCE~\cite{milnce}                        & 34.2            & 30.7           \\
        InternVideo-MM-L14~\cite{internvideo}             & 40.1            & 37.3           \\
        TAN~\cite{tan}                            & 49.4            & -              \\
        TAN* (Joint, S1, PE+LC)~\cite{vina} & 63              & 31.2           \\
        VINA(w/o nar)~\cite{vina}                    & -               & 35.6           \\
        VINA~\cite{vina}                   & 66.5            & 37.4           \\ \midrule
        Ours                           & \textbf{69.3}   & \textbf{43.3}  \\ \bottomrule
    \end{tabular}
    \caption{Comparison with state-of-the-art methods for the narration grounding and step grounding on the HTM-Align and HT-Step datasets, respectively.}
\label{tbl:sota_htstep}
\vspace{-10pt}
\end{table}

\begin{table}[t]
    \centering
    \footnotesize
    \setlength{\tabcolsep}{5mm}
    \setlength{\extrarowheight}{0.2mm}
    \begin{tabular}{@{}cc@{}}
        \toprule
        Method           & ↑Avg. R@1 (\%)        \\ \midrule
        HT100M~\cite{howto100m}           & 33.6                 \\
        VideoCLIP~\cite{xu2021videoclip}        & 33.9                 \\
        MCN~\cite{mcn}              & 35.1                 \\
        DWSA~\cite{dwsa}             & 35.3                 \\
        MIL-NCE~\cite{milnce}        & 40.5                 \\
        Zhukov~\cite{crosstask}           & 40.5                 \\
        VT-TWINS*~\cite{ko2022video}        & 40.7                 \\
        UniVL~\cite{luo2020univl}            & 42.0                   \\
        VINA~\cite{vina}             & 44.8                 \\ \midrule
        Ours    & \textbf{47.9}                 \\\bottomrule
        %ours (Nar + LLM) & 45.9 \\ 
    \end{tabular}
    \caption{Comparison with state-of-the-art methods for action step localization on CrossTask Dataset.}
    \label{tbl:al}    
    \vspace{-10pt}
\end{table}

\subsection{Comparison with State-of-the-Art Approaches}

\subsubsection{Results For Narration and Step Grounding} We compare our method with state-of-the-art approaches for narration and step grounding tasks on the HTM-Align and HT-Step datasets, and the results are presented in Table~\ref{tbl:sota_htstep}. The results for narration grounding on the HTM-Align dataset were obtained by training a model that adds positional embedding to the text input, following previous work~\cite{vina}.  We find that our method significantly outperforms previous approaches, achieving state-of-the-art performance for both narration and step-grounding tasks. Notably, our method improves the performance of VINA, which uses narrations and procedure steps from the wikiHow dataset for training, by 2.8\% and 5.9\% for narration and step grounding, respectively. Moreover, VINA~\cite{vina} additionally uses narrations of videos during test time. In a fair comparison, where narrations are not available during testing, our method outperforms VINA (VINA w/o nar) by 7.7\%. The results further demonstrate the effectiveness of our proposed methods.

\subsubsection{Results for Action Step Localization} We further compare our model with other works on action step localization using the CrossTask dataset. The results are reported in Table \ref{tbl:al}. Our method improves the state-of-the-art performance by 3.1\%. The results show the strong transfer capabilities of our learned models to different downstream datasets. In addition, the results indicate that our model can handle text with varying levels of information granularity, from abstract action steps to detailed, enriched procedure steps and various narrations.

\section{Conclusions} 
In this work, we introduce a novel learning framework designed to train models for localizing procedure steps in noisy, narrated instructional videos. Initially, we utilize LLMs to filter out text information irrelevant to the tasks depicted in the videos and to summarize task-related procedural steps from the narrations. Subsequently, we employ our proposed Multi-Pathway Text-Video Alignment strategy to generate pseudo-alignments between LLM-steps and video segments for training purposes. Extensive experiments conducted across three datasets for three downstream tasks show that our method establishes new state-of-the-art benchmarks. Furthermore, ablation studies confirm that the procedural steps extracted by our framework serve as superior text inputs compared to both narration and procedural steps derived from human-constructed knowledge bases.

\bigskip
\noindent\textbf{Acknowledgements} This research has been partially funded by research grants to Dimitris N. Metaxas through NSF: 2310966, 2235405, 2212301, 2003874, and FA9550-23-1-0417 and NIH 2R01HL127661.

\clearpage

\bibliographystyle{splncs04}
\bibliography{main}

\begin{thebibliography}{10}
\providecommand{\url}[1]{\texttt{#1}}
\providecommand{\urlprefix}{URL }
\providecommand{\doi}[1]{https://doi.org/#1}

\bibitem{ahn2022can}
Ahn, M., Brohan, A., Brown, N., Chebotar, Y., Cortes, O., David, B., Finn, C., Fu, C., Gopalakrishnan, K., Hausman, K., et~al.: Do as i can, not as i say: Grounding language in robotic affordances. arXiv preprint arXiv:2204.01691  (2022)

\bibitem{bain2021frozen}
Bain, M., Nagrani, A., Varol, G., Zisserman, A.: Frozen in time: A joint video and image encoder for end-to-end retrieval. In: Proceedings of the IEEE/CVF International Conference on Computer Vision. pp. 1728--1738 (2021)

\bibitem{brown2020language}
Brown, T., Mann, B., Ryder, N., Subbiah, M., Kaplan, J.D., Dhariwal, P., Neelakantan, A., Shyam, P., Sastry, G., Askell, A., et~al.: Language models are few-shot learners. Advances in neural information processing systems  \textbf{33},  1877--1901 (2020)

\bibitem{activitynet}
Caba~Heilbron, F., Escorcia, V., Ghanem, B., Carlos~Niebles, J.: Activitynet: A large-scale video benchmark for human activity understanding. In: Proceedings of the ieee conference on computer vision and pattern recognition. pp. 961--970 (2015)

\bibitem{cao2022locvtp}
Cao, M., Yang, T., Weng, J., Zhang, C., Wang, J., Zou, Y.: Locvtp: Video-text pre-training for temporal localization. In: European Conference on Computer Vision. pp. 38--56. Springer (2022)

\bibitem{k700}
Carreira, J., Zisserman, A.: Quo vadis, action recognition? a new model and the kinetics dataset. In: proceedings of the IEEE Conference on Computer Vision and Pattern Recognition. pp. 6299--6308 (2017)

\bibitem{chang2019d3tw}
Chang, C.Y., Huang, D.A., Sui, Y., Fei-Fei, L., Niebles, J.C.: D3tw: Discriminative differentiable dynamic time warping for weakly supervised action alignment and segmentation. In: Proceedings of the IEEE/CVF Conference on Computer Vision and Pattern Recognition. pp. 3546--3555 (2019)

\bibitem{mcn}
Chen, B., Rouditchenko, A., Duarte, K., Kuehne, H., Thomas, S., Boggust, A., Panda, R., Kingsbury, B., Feris, R., Harwath, D., et~al.: Multimodal clustering networks for self-supervised learning from unlabeled videos. In: Proceedings of the IEEE/CVF International Conference on Computer Vision. pp. 8012--8021 (2021)

\bibitem{chen2023visual}
Chen, D., Liu, J., Dai, W., Wang, B.: Visual instruction tuning with polite flamingo. arXiv preprint arXiv:2307.01003  (2023)

\bibitem{wsag}
Chen, L., Niu, Y., Chen, B., Lin, X., Han, G., Thomas, C., Ayyubi, H., Ji, H., Chang, S.F.: Weakly-supervised temporal article grounding. arXiv preprint arXiv:2210.12444  (2022)

\bibitem{fdt}
Chen, Y., Yuan, J., Tian, Y., Geng, S., Li, X., Zhou, D., Metaxas, D.N., Yang, H.: Revisiting multimodal representation in contrastive learning: from patch and token embeddings to finite discrete tokens. In: Proceedings of the IEEE/CVF Conference on Computer Vision and Pattern Recognition. pp. 15095--15104 (2023)

\bibitem{chen2022hierarchically}
Chen, Y., Zhao, L., Yuan, J., Tian, Y., Xia, Z., Geng, S., Han, L., Metaxas, D.N.: Hierarchically self-supervised transformer for human skeleton representation learning. In: European Conference on Computer Vision. pp. 185--202. Springer (2022)

\bibitem{instructblip}
Dai, W., Li, J., Li, D., Tiong, A.M.H., Zhao, J., Wang, W., Li, B., Fung, P., Hoi, S.C.H.: Instructblip: Towards general-purpose vision-language models with instruction tuning. In: Oh, A., Naumann, T., Globerson, A., Saenko, K., Hardt, M., Levine, S. (eds.) Advances in Neural Information Processing Systems 36: Annual Conference on Neural Information Processing Systems 2023, NeurIPS 2023, New Orleans, LA, USA, December 10 - 16, 2023 (2023), \url{http://papers.nips.cc/paper\_files/paper/2023/hash/9a6a435e75419a836fe47ab6793623e6-Abstract-Conference.html}

\bibitem{driess2023palm}
Driess, D., Xia, F., Sajjadi, M.S., Lynch, C., Chowdhery, A., Ichter, B., Wahid, A., Tompson, J., Vuong, Q., Yu, T., et~al.: Palm-e: An embodied multimodal language model. arXiv preprint arXiv:2303.03378  (2023)

\bibitem{dropdtw}
Dvornik, M., Hadji, I., Derpanis, K.G., Garg, A., Jepson, A.: Drop-dtw: Aligning common signal between sequences while dropping outliers. Advances in Neural Information Processing Systems  \textbf{34},  13782--13793 (2021)

\bibitem{stepformer}
Dvornik, N., Hadji, I., Zhang, R., Derpanis, K.G., Wildes, R.P., Jepson, A.D.: Stepformer: Self-supervised step discovery and localization in instructional videos. In: Proceedings of the IEEE/CVF Conference on Computer Vision and Pattern Recognition. pp. 18952--18961 (2023)

\bibitem{procel}
Elhamifar, E., Naing, Z.: Unsupervised procedure learning via joint dynamic summarization. In: Proceedings of the IEEE/CVF International Conference on Computer Vision. pp. 6341--6350 (2019)

\bibitem{crosstask}
Gan, Z., Li, L., Li, C., Wang, L., Liu, Z., Gao, J., et~al.: Vision-language pre-training: Basics, recent advances, and future trends. Foundations and Trends{\textregistered} in Computer Graphics and Vision  \textbf{14}(3--4),  163--352 (2022)

\bibitem{hiclip}
Geng, S., Yuan, J., Tian, Y., Chen, Y., Zhang, Y.: Hiclip: Contrastive language-image pretraining with hierarchy-aware attention. arXiv preprint arXiv:2303.02995  (2023)

\bibitem{chatgptann}
Gilardi, F., Alizadeh, M., Kubli, M.: Chatgpt outperforms crowd-workers for text-annotation tasks. arXiv preprint arXiv:2303.15056  (2023)

\bibitem{tan}
Han, T., Xie, W., Zisserman, A.: Temporal alignment networks for long-term video. In: Proceedings of the IEEE/CVF Conference on Computer Vision and Pattern Recognition. pp. 2906--2916 (2022)

\bibitem{jia2021scaling}
Jia, C., Yang, Y., Xia, Y., Chen, Y.T., Parekh, Z., Pham, H., Le, Q., Sung, Y.H., Li, Z., Duerig, T.: Scaling up visual and vision-language representation learning with noisy text supervision. In: International conference on machine learning. pp. 4904--4916. PMLR (2021)

\bibitem{kinetics}
Kay, W., Carreira, J., Simonyan, K., Zhang, B., Hillier, C., Vijayanarasimhan, S., Viola, F., Green, T., Back, T., Natsev, P., et~al.: The kinetics human action video dataset. arXiv preprint arXiv:1705.06950  (2017)

\bibitem{ko2022video}
Ko, D., Choi, J., Ko, J., Noh, S., On, K.W., Kim, E.S., Kim, H.J.: Video-text representation learning via differentiable weak temporal alignment. In: Proceedings of the IEEE/CVF Conference on Computer Vision and Pattern Recognition. pp. 5016--5025 (2022)

\bibitem{wikihow}
Koupaee, M., Wang, W.Y.: Wikihow: A large scale text summarization dataset. arXiv preprint arXiv:1810.09305  (2018)

\bibitem{lea2016temporal}
Lea, C., Vidal, R., Reiter, A., Hager, G.D.: Temporal convolutional networks: A unified approach to action segmentation. In: Computer Vision--ECCV 2016 Workshops: Amsterdam, The Netherlands, October 8-10 and 15-16, 2016, Proceedings, Part III 14. pp. 47--54. Springer (2016)

\bibitem{maxi}
Lin, W., Karlinsky, L., Shvetsova, N., Possegger, H., Kozinski, M., Panda, R., Feris, R., Kuehne, H., Bischof, H.: Match, expand and improve: Unsupervised finetuning for zero-shot action recognition with language knowledge. arXiv preprint arXiv:2303.08914  (2023)

\bibitem{distsup}
Lin, X., Petroni, F., Bertasius, G., Rohrbach, M., Chang, S.F., Torresani, L.: Learning to recognize procedural activities with distant supervision. In: Proceedings of the IEEE/CVF Conference on Computer Vision and Pattern Recognition. pp. 13853--13863 (2022)

\bibitem{liu2023improved}
Liu, H., Li, C., Li, Y., Lee, Y.J.: Improved baselines with visual instruction tuning. arXiv preprint arXiv:2310.03744  (2023)

\bibitem{llava}
Liu, H., Li, C., Wu, Q., Lee, Y.J.: Visual instruction tuning. Advances in neural information processing systems  \textbf{36} (2024)

\bibitem{adamw}
Loshchilov, I., Hutter, F.: Decoupled weight decay regularization. arXiv preprint arXiv:1711.05101  (2017)

\bibitem{setsup}
Lu, Z., Elhamifar, E.: Set-supervised action learning in procedural task videos via pairwise order consistency. In: Proceedings of the IEEE/CVF Conference on Computer Vision and Pattern Recognition. pp. 19903--19913 (2022)

\bibitem{luo2020univl}
Luo, H., Ji, L., Shi, B., Huang, H., Duan, N., Li, T., Li, J., Bharti, T., Zhou, M.: Univl: A unified video and language pre-training model for multimodal understanding and generation. arXiv preprint arXiv:2002.06353  (2020)

\bibitem{vina}
Mavroudi, E., Afouras, T., Torresani, L.: Learning to ground instructional articles in videos through narrations. arXiv preprint arXiv:2306.03802  (2023)

\bibitem{milnce}
Miech, A., Alayrac, J.B., Smaira, L., Laptev, I., Sivic, J., Zisserman, A.: End-to-end learning of visual representations from uncurated instructional videos. In: Proceedings of the IEEE/CVF Conference on Computer Vision and Pattern Recognition. pp. 9879--9889 (2020)

\bibitem{howto100m}
Miech, A., Zhukov, D., Alayrac, J.B., Tapaswi, M., Laptev, I., Sivic, J.: Howto100m: Learning a text-video embedding by watching hundred million narrated video clips. In: Proceedings of the IEEE/CVF international conference on computer vision. pp. 2630--2640 (2019)

\bibitem{pu2023summarization}
Pu, X., Gao, M., Wan, X.: Summarization is (almost) dead. arXiv preprint arXiv:2309.09558  (2023)

\bibitem{clip}
Radford, A., Kim, J.W., Hallacy, C., Ramesh, A., Goh, G., Agarwal, S., Sastry, G., Askell, A., Mishkin, P., Clark, J., et~al.: Learning transferable visual models from natural language supervision. In: International conference on machine learning. pp. 8748--8763. PMLR (2021)

\bibitem{t5}
Raffel, C., Shazeer, N., Roberts, A., Lee, K., Narang, S., Matena, M., Zhou, Y., Li, W., Liu, P.J.: Exploring the limits of transfer learning with a unified text-to-text transformer. The Journal of Machine Learning Research  \textbf{21}(1),  5485--5551 (2020)

\bibitem{dtw}
Sakoe, H., Chiba, S.: Dynamic programming algorithm optimization for spoken word recognition. IEEE transactions on acoustics, speech, and signal processing  \textbf{26}(1),  43--49 (1978)

\bibitem{dwsa}
Shen, Y., Wang, L., Elhamifar, E.: Learning to segment actions from visual and language instructions via differentiable weak sequence alignment. In: Proceedings of the IEEE/CVF Conference on Computer Vision and Pattern Recognition. pp. 10156--10165 (2021)

\bibitem{howtocap}
Shvetsova, N., Kukleva, A., Hong, X., Rupprecht, C., Schiele, B., Kuehne, H.: Howtocaption: Prompting llms to transform video annotations at scale. arXiv preprint arXiv:2310.04900  (2023)

\bibitem{llmann}
Tan, Z., Beigi, A., Wang, S., Guo, R., Bhattacharjee, A., Jiang, B., Karami, M., Li, J., Cheng, L., Liu, H.: Large language models for data annotation: A survey. arXiv preprint arXiv:2402.13446  (2024)

\bibitem{coin}
Tang, Y., Ding, D., Rao, Y., Zheng, Y., Zhang, D., Zhao, L., Lu, J., Zhou, J.: Coin: A large-scale dataset for comprehensive instructional video analysis. In: Proceedings of the IEEE/CVF Conference on Computer Vision and Pattern Recognition. pp. 1207--1216 (2019)

\bibitem{llamav1}
Touvron, H., Lavril, T., Izacard, G., Martinet, X., Lachaux, M.A., Lacroix, T., Rozi{\`e}re, B., Goyal, N., Hambro, E., Azhar, F., et~al.: Llama: Open and efficient foundation language models. arXiv preprint arXiv:2302.13971  (2023)

\bibitem{llamav2}
Touvron, H., Martin, L., Stone, K., Albert, P., Almahairi, A., Babaei, Y., Bashlykov, N., Batra, S., Bhargava, P., Bhosale, S., et~al.: Llama 2: Open foundation and fine-tuned chat models. arXiv preprint arXiv:2307.09288  (2023)

\bibitem{van2023clinical}
Van~Veen, D., Van~Uden, C., Blankemeier, L., Delbrouck, J.B., Aali, A., Bluethgen, C., Pareek, A., Polacin, M., Reis, E.P., Seehofnerova, A., et~al.: Clinical text summarization: Adapting large language models can outperform human experts. Research Square  (2023)

\bibitem{selfatt}
Vaswani, A., Shazeer, N., Parmar, N., Uszkoreit, J., Jones, L., Gomez, A.N., Kaiser, {\L}., Polosukhin, I.: Attention is all you need. Advances in neural information processing systems  \textbf{30} (2017)

\bibitem{internvideo}
Wang, Y., Li, K., Li, Y., He, Y., Huang, B., Zhao, Z., Zhang, H., Xu, J., Liu, Y., Wang, Z., et~al.: Internvideo: General video foundation models via generative and discriminative learning. arXiv preprint arXiv:2212.03191  (2022)

\bibitem{xu2021videoclip}
Xu, H., Ghosh, G., Huang, P.Y., Okhonko, D., Aghajanyan, A., Metze, F., Zettlemoyer, L., Feichtenhofer, C.: Videoclip: Contrastive pre-training for zero-shot video-text understanding. arXiv preprint arXiv:2109.14084  (2021)

\bibitem{clipvip}
Xue, H., Sun, Y., Liu, B., Fu, J., Song, R., Li, H., Luo, J.: Clip-vip: Adapting pre-trained image-text model to video-language representation alignment. arXiv preprint arXiv:2209.06430  (2022)

\bibitem{llmsent}
Ye, J., Gao, J., Li, Q., Xu, H., Feng, J., Wu, Z., Yu, T., Kong, L.: Zerogen: Efficient zero-shot learning via dataset generation. arXiv preprint arXiv:2202.07922  (2022)

\bibitem{zhao2024videoprism}
Zhao, L., Gundavarapu, N.B., Yuan, L., Zhou, H., Yan, S., Sun, J.J., Friedman, L., Qian, R., Weyand, T., Zhao, Y., et~al.: Videoprism: A foundational visual encoder for video understanding. arXiv preprint arXiv:2402.13217  (2024)

\bibitem{antgpt}
Zhao, Q., Zhang, C., Wang, S., Fu, C., Agarwal, N., Lee, K., Sun, C.: Antgpt: Can large language models help long-term action anticipation from videos? arXiv preprint arXiv:2307.16368  (2023)

\bibitem{zhou2023procedure}
Zhou, H., Mart{\'\i}n-Mart{\'\i}n, R., Kapadia, M., Savarese, S., Niebles, J.C.: Procedure-aware pretraining for instructional video understanding. In: Proceedings of the IEEE/CVF Conference on Computer Vision and Pattern Recognition. pp. 10727--10738 (2023)

\end{thebibliography}

\clearpage
\appendix

\title{Supplementary Material}

\titlerunning{Localizing Actions in Videos with LLM-Based Multi-Pathway Alignment}

\author{Yuxiao Chen\inst{1} \and
Kai Li \inst{2} \and
Wentao Bao\inst{4} \and Deep Patel \inst{3} \and \\ Yu Kong \inst{4} \and Martin Renqiang Min \inst{3} \and Dimitris N. Metaxas \inst{1}}

\authorrunning{Y. Chen et al.}

\institute{Rutgers University \email{\{yc984, dnm\}@cs.rutgers.edu}, \and
Meta \email{\{li.gml.kai@gmail.com\}} \and
NEC Labs America-Princeton \email{\{dpatel, renqiang\}@nec-labs.com}
 \and
Michigan State University
\email{\{baowenta, yukong\}@msu.edu}}

\maketitle

\section{Implementation Details}

\subsection{Extracting LLM-Steps} We apply the Llama 2 7B model \cite{llamav2} to extract LLM-steps from narrations. The sentencified narrations provided by Lin \etal \cite{tan} serve as the input. We divide the narrations of each video into non-overlapped segments, each consisting of 10 narration sentences. The segments are then individually processed by the LLM.

\subsection{Multi-Pathway Text-Video Alignment}

For the S-N-V pathway, we apply the text encoder from the \textit{TAN* (Joint, S1, LC)} model, which is implemented and trained following Mavroudi \etal~\cite{vina}, as $E_t$ to extract the feature representations of narrations and LLM-steps. In the S-N-Long pathway, the text encoder and video encoder of \textit{TAN* (Joint, S1, LC)} are used as $E_t^L$ and $E_v^L$, respectively. For the  S-N-short pathways, we set $E_t^S$ and $E_v^S$ as the pre-trained text and video encoders of the InternVideo-MM-L14 model~\cite{internvideo}.

\subsection{Model Architecture} Following previous works~\cite{vina,tan}, we apply the S3D model and Bag-of-Words approach based on Word2Vec embeddings released by Miech \etal  ~\cite{milnce} as the video backbone ($f_{b}$) and  text encoder ($g_{b}$), respectively. we keep the video backbone frozen during the training following~\cite{vina,tan}. The extracted video segment or LLM-step embeddings from these models are projected onto a 256-dimensional space using a fully connected layer before being fed into the video or text Transformer. The learnable positional embeddings~\cite{tan} are added with video segment features to incorporate the temporal order information. The video Transformer ($f_{trs}$) consists of two Transformer encoder layers~\cite{selfatt}, each with 8 attention heads and a 256-dimensional representation. The text Transformer  $g_{trs}$ and Joint transformer $h_{trs}$ are configured with the same settings as the video Transformer.

\subsection{Model Training} Following the common practice~\cite{tan,vina}, we divide each input video into a sequence of 1-second non-overlapped segments and set the frame rate to 16 FPS. The video backbone, $f_b$~\cite{milnce}, is used to extract a feature embedding from each segment. The maximal input video segment length is set to 1024. We set the batch size to 32. We apply the AdamW optimizer~\cite{adamw} with a learning rate of $2 \times 10^{-4}$ and weight decay ratio $1 \times 10^{-5}$ for training. We train the models for 12 epochs. The learning rate is linearly warmed up for the first 1000 iterations and followed by a cosine decay.

\subsection{Downstream Tasks}
Following the common setting~\cite{vina}, the trained models are directly tested on the downstream datasets without additional fine-tuning. The trained models take the text descriptions of procedural steps (action steps or narrations) and video segments as input, and output the similarity scores between procedural steps and video segments. The metric for each downstream task is calculated from these similarity scores.

\section{Additional Experiment Results}
\subsection{Comparison of different types of text information} 
\begin{figure}[t]
\centering
\includegraphics[width=\linewidth]{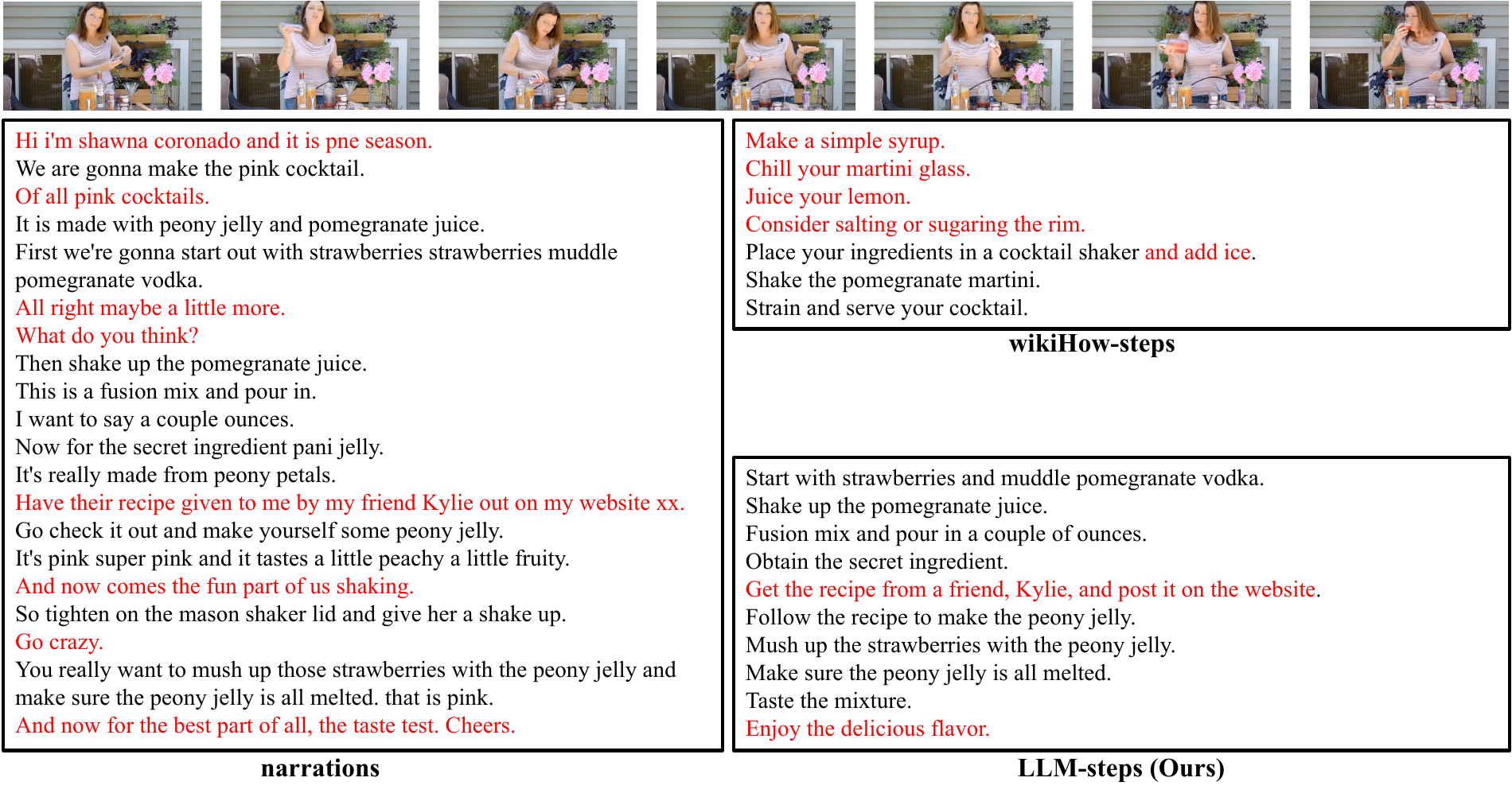}
  \caption{Comparison of different types of text information associated with the instruction video. The sentences highlighted in red are irrelevant to the tasks demonstrated in the video.}
\label{fig:sup_fig_1}
\end{figure}

\begin{figure}[t]
\centering
\includegraphics[width=\linewidth]{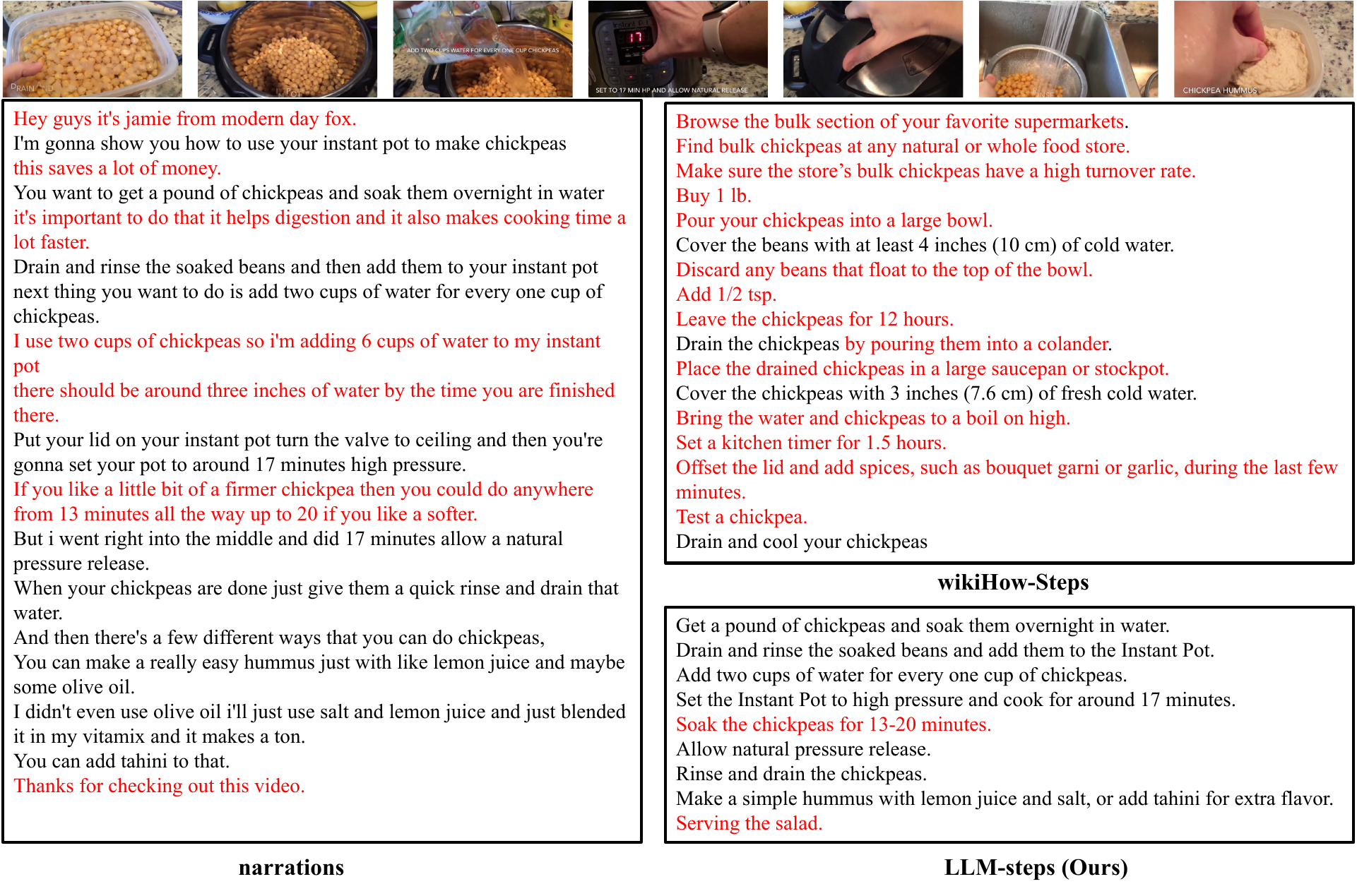}
  \caption{Comparison of different types of text information associated with the instruction video. The sentences highlighted in red are irrelevant to the tasks demonstrated in the video.}
\label{fig:sup_fig_2}
\end{figure}

\begin{figure}[t]
\centering
\includegraphics[width=\linewidth]{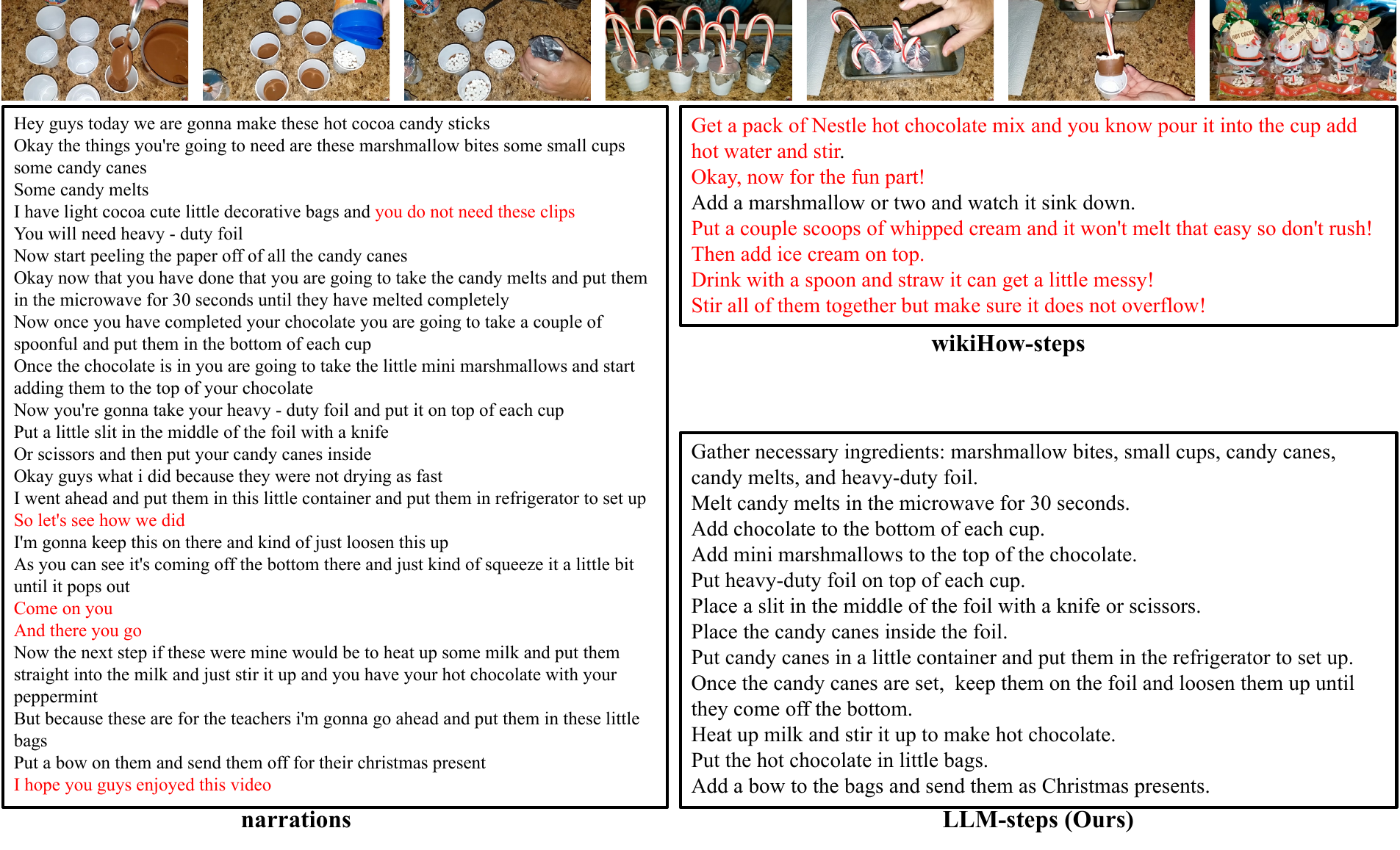}
  \caption{Comparison of different types of text information associated with the instruction video. The sentences highlighted in red are irrelevant to the tasks demonstrated in the video.}
\label{fig:sup_fig_3}
\end{figure}

We additionally show three videos and their narrations, wikiHow-Steps extracted following~\cite{tan}, and LLM-steps in Figure~\ref{fig:sup_fig_1}, \ref{fig:sup_fig_2}, and~\ref{fig:sup_fig_3}. These examples further demonstrate that our LLM steps are more descriptive and contain less information that is unrelated to the video task than narrations. In addition, we observe that LLM-steps align the task contents better than wikiHow-steps, demonstrating the effectiveness of our LLM-steps.

\clearpage


\end{document}